\documentclass[wcp]{jmlr}

\usepackage{longtable}% for long tables
\usepackage{booktabs}
\usepackage{algorithm}
\usepackage{algorithmic}
\usepackage{multirow}
\usepackage{amsmath}
\usepackage{amssymb}
\usepackage{xcolor}
\usepackage{bm}

\usepackage{graphics}
\usepackage{graphicx}
\jmlrvolume{60}
\jmlryear{2016}
\jmlrworkshop{ACML 2016}
\title[Short Title]{hi-RF: Incremental Learning Random Forest for large-scale multi-class Data Classification}
  \author{\Name{Tingting Xie} \Email{xietingting14@nudt.edu.cn}\\
  \Name{Yuxing Peng} \Email{pengyuxing@aliyun.com}\\
  \Name{Changjian Wang} \Email{c\_j\_wang@yeah.com}\\
  \addr National Lab for Parallel and Distributed Processing, School of Computer, National University of Defense Technology, China, 410073
  }
\begin{document}
\maketitle
% Abstract part
\begin{abstract}
In recent years, dynamically growing data and incrementally growing number of classes pose new challenges to large-scale data classification research. Most traditional methods struggle to balance the precision and computational burden when data and its number of classes increased. However, some methods are with weak precision, and the others are time-consuming. In this paper, we propose an incremental learning method, namely, heterogeneous incremental Nearest Class Mean Random Forest (hi-RF), to handle this issue. It is a heterogeneous method that either replaces trees or updates trees leaves in the random forest adaptively, to reduce the computational time in comparable performance, when data of new classes arrive. Specifically, to keep the accuracy, one proportion of trees are replaced by new NCM decision trees; to reduce the computational load, the rest trees are updated their leaves probabilities only. Most of all, out-of-bag estimation and out-of-bag boosting are proposed to balance the accuracy and the computational efficiency. Fair experiments were conducted and demonstrated its comparable precision with much less computational time.
\end{abstract}
\begin{keywords}
large scale multi-class classification; Incremental Learning; random forest; heterogeneous incremental Nearest Class Mean Random Forest
\end{keywords}

% Introduction part
\section{Introduction}
With data increasingly available in every second, automatic classification has attracted wide attention in both research and industry. Though there are thousands of classes in this dataset, new visual classes and data grow dynamically in practice. To retrain a model from scratch when new data arrives is very time-consuming and requires full access to the original training data.\par
Among the state of the art solutions, Random Forest (RF)~\citep{ho1995random}, \citep{ho1998random} has been tested as a successful representative for large-scale image classification given its performance of high efficiency and accuracy. As a standard supervised learning method, RF training usually assumes that the number of class is fixed and the class distribution is fixed~\citep{xu2012improved}, which is unable to handle dynamic growing data and classes. On-line variants of RF~\citep{basak2004online}, \citep{utgoff1997decision} were proposed to address one aspect of this issue-the class distribution is fixed. They assume the number of classes as well as the class labels are known beforehand, so they cannot handle new classes of data.\par

% Figure 1
\begin{figure}
\centering
\includegraphics[width=0.8\textwidth]{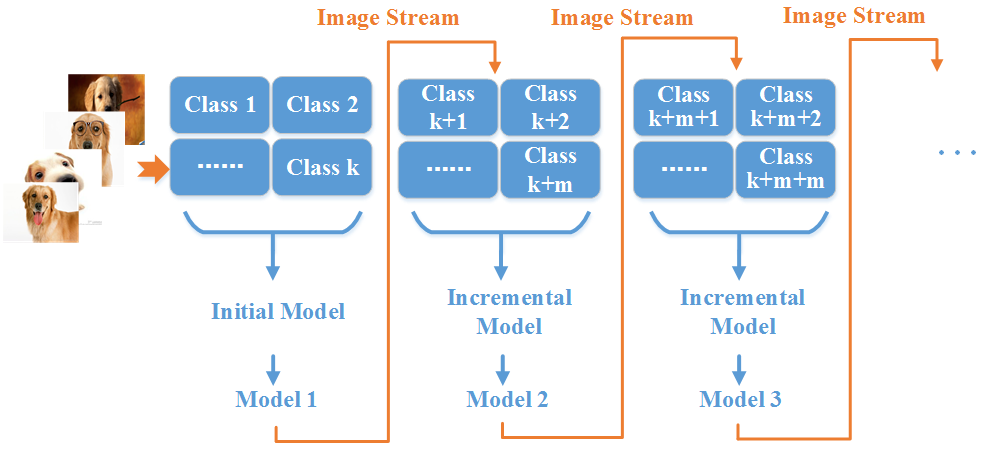}
\caption{Incremental learning concept.}\label{fig:figure1}
\vspace{-0.5cm}
\end{figure}
The problem, that a few classes are available at the beginning and new classes of data arrive consequentially, is called incremental learning problem. The conceptual structure of the incremental learning is illustrated in Figure~\ref{fig:figure1}. Previous methods often re-use subtrees in incremental learning and only handle data from one new class. In order to achieve high accuracy, they have to re-use a large proportion of subtrees, which is very time-consuming. Their stiff procedure (i.e., they can only handle the arriving data of one class) and high computational load limit the generalization. To deal with these problems, we propose a large-scale multi-class data classification method-Heterogeneous Incremental Nearest Class Mean Random Forest (hi-RF). The method can handle multiple new classes of data and reduce computational cost in comparable accuracy.\par
In hi-RF, Rolling Release NCM decision trees (RRN) was presented to integrate new classes. For a proportion of trees in RF, RRN retrained them with new NCM decision trees~\citep{ristin2014incremental}, based on their out-of-bag error~\citep{breiman1996out} comparing with a particular threshold, once fresh data arrives. The training dataset is the subset of the new and the old training samples for each NCM decision tree, so it is convenient to add upcoming samples and keep the accuracy. Because of that, we can address the issue of multiple new classes. To reduce the computational cost, Regenerate leaves probabilities (RLP) was advanced for the rest proportion of trees in RF. RLP only updates the probabilities of leaf node, but not modifies any part of the decision tree. Its negligible time cost reduces the computing load greatly. To balance the accuracy and computational cost, it is very important to select a particular threshold to separate RF into two parts. The threshold selection is based on Out-Of-Bag estimation (OOB estimation), which is calculated by out-of-bag error. Most time, we usually limit the proportion of re-training by decreasing the threshold as small as possible to alleviate the computational burden with comparable accuracy. During the procedure of rolling release new data, to make OOB estimation balance the trees better and more stability to new data, Out-Of-Bag boosting (OOB boosting) is proposed. OOB boosting increases the out-of-bag error of trees updated by RLP, so there is more chance to retrain the trees as the next time new data added, and it preserves the stability of trees and maintains the accuracy. \par
In this work, the contribution of the paper are as follows. First, RRN was adopted to retrain the decision tree based on their out-of-bag error in a rolling manner, thus providing freedom to integrate multiple class of data. Second, RLP was presented to reduce the computational load. As a sequence, it performs pretty good on the challenging large-scale ImageNet datasets, because the computational load is a little more than half of that for re-training all the trees. Experiments showed that, compared with off-line random forest, the computational time of hi-RF is a little more than half of the off-line, with the accuracy loss $2.70\%$. Third, OOB estimation is used to balance computational load and accuracy by adaptively cutting down the proper proportion of decision trees, re-training trees and updating trees. Besides, OOB boosting helps a lot to maintain the trees stability in the incremental procedure. \par
This paper is organized as follows: we firstly summarize related works in Section~\ref{related work}. Section~\ref{hi-RF} demonstrates the whole working procedure of hi-RF. Experiments and brief discussion will be presented in Section~\ref{experiment} and we will conclude with a summary in Section~\ref{conclusion}.

% Related work part
\section{Related work}
\label{related work}
Since ImageNet appeared several years ago, many researchers are working on data classification with RF. Compared to other classifiers, RF have performed very well in classification, visual tracking~\citep{saffari2009line}, feature detection~\citep{yang2013privileged}, and even in cancer detection from mass proteomic pictures. RF have been a good candidate for computer vision for several reasons. Firstly, they run fast both in training and classification. Secondly, they can be easily parallelized, thus they are always considered for distributed computing and GPU acceleration. Additionally, RF have inherently hierarchical structure, so they can be made locally modification in deep layers, which only affect part of the data in negligible cost~\citep{ristin2014incremental}, \citep{ristin2016incremental}. Above all, compared to SVM and other ensemble methods, RF are naturally multi-class, more robust against noises and more suitable for generalization~\citep{gall2011hough}, \citep{schulter2011line}.\par
Researchers has proposed dimensionality reduction methods to address fine-grained classification. \citep{xu2012improved} is a particular one, and it effectively reduce subspace size as well as improve classification performance. \citep{tang2012efficient} adopts sparse coding to train a dictionary of visual words and then convert SIFT descriptors into sparse vectors. However, the methods above were applied to off-line learning, so they cannot process data grown continuously.
Many papers used RF in on-line mode to address the problem of large data. They often combine on-line bagging and on-line decision trees with random feature selection, but they are either memory intensive because of storing all the data in every node \citep{utgoff1997decision} or have to discard important information if parent nodes change. Besides, there is some researchers focus on improving the accuracy. Because of the hierarchy structure, error can be propagated further down to the tree. While some methods solve this problem, they combine decision trees with ideas from neural networks~\citep{basak2004online}, but they often lose the $O(log n)$ evaluation time because samples are propagated to all nodes. \citep{saffari2009line} proposed a novel on-line algorithm that has neither of the problems. The algorithm allows discarding entire trees by out-of-bag-error and consecutively growing of new trees. Nevertheless, in practice, data streams might obtain new classes, and they cannot adapt to this problem.\par
In incremental learning, algorithms was established with competitive accuracy to off-line RF without retraining the whole forests. The approach~\citep{mensink2013distance} learns a discriminative metric on the initial set of classes, and classifies samples simply based on the nearest class mean. Adding a new class means inserting its mean in the pool of classes, leading to a negligible computational burden. They could not update the structure of forests by itself when a new class added, which will finally lead to suboptimal performance. Conversely, we propose to update the structure of hi-RF to integrate new classes. While in~\citep{ristin2014incremental}, \citep{ristin2016incremental}, they proposed a novel and almost perfect method to solve the problem, but they can only process data of one class added. That is to say, it is hard to process new classes in batch, which limits their usability.\par
Transfer learning is also along with incremental learning, while it intends to reduce the amount of labeled data required to learn a new class~\citep{gao2014transfer}. Incremental learning is different from it for two reasons. Firstly, transfer learning is limited to one-vs-all classification, Secondly, the intention of incremental learning is to add a new class efficiently instead of exploiting the knowledge from previous classes to reduce the amount of annotation necessary for good performance.\par

In this paper, we demonstrate a heterogeneous approach, inspired by on-line bagging and nearest class mean. Usually, there are trees containing too much noise, and they hamper the accuracy. It is necessary to replace trees in poor performance. And the simple use of class means as centroids makes it efficient to train a decision tree, and it results in high performance with much less cost. Finally, hi-RF is proposed to achieve better performance.

% Model part
\section{Heterogeneous incremental nearest class mean random forest}
\label{hi-RF}
Random Forest (RF) is an ensemble classifier that aggregates many decision trees. It does a popularity vote of individual tree when predicting a class. This character makes RF suitable to incremental learning. \par
Heterogeneous incremental Nearest Class Mean Random Forest (hi-RF) is a new way to modify the random forest for incrementally integrated data. It is constructed by Nearest Class Mean decision trees (NCM decision trees)~\citep{ristin2014incremental}, \citep{ristin2016incremental}. When new data arrives, hi-RF processes the trees in the forest in two ways according a standard, which is calculated by OOB estimation and OOB boosting, which will be described in~\ref{oob}. For each tree, if it does not satisfy the standard, it will be retrained with bootstrap samples which combined previous and fresh data. Otherwise, it will be updated with the leaves node probabilities changed only. The former procedure of retraining is called RRN and the latter of updating is named RLP. The whole process is depicted in Algorithm~\ref{alg:hi-RF}, especially, RRN can be presented by line 2:4, and RLP can be presented by 2,5:6, which two will be described in section~\ref{alg:RRN} and~\ref{alg:RLP}. \par
% Algorithm 1:hi-RF framework
\begin{algorithm}[htb]
\caption{Heterogeneous incremental Nearest Class Mean Random Forest (hi-RF)}
\label{alg:hi-RF}
\begin{algorithmic}[1]

\REQUIRE~~
\\
Previous model, $m$; \\
Number of decision trees in $m$, $s$; \\
The set of out-of-bag error for each tree, $O$; \\
Old training data, $D_o$; \\
New training data, $D_n$; \\
\ENSURE~~
\\
New model, $M$;
\FOR{each time new data arriving}
	\STATE threshold $\leftarrow OOB\_estimation(O)$
	\FOR{each tree $T_i$ in $m$}
	\IF{$T_i$ does not reach the threshold}
	\STATE $T_i \leftarrow Retraining(D^o,D^n)$
	\ELSE
	\STATE $T_i \leftarrow Updating(D^o,D^n)$
	\ENDIF
	\ENDFOR
	\STATE $O\leftarrow OOB\_boosting(O)$
	\STATE $M \leftarrow Bagging(T_1, T_2,...,T_s)$
\ENDFOR

\RETURN $M$
\end{algorithmic}
\end{algorithm}
% Section Variable descriptions
To make mathematical model understood easier, we represent variable description in Table~\ref{tab:variable_description}. hi-RF is aggregated by $s$ versions of decision tree, and each tree is denoted by $T_i (i \in \left\{1,2,...,s\right\})$~\citep{breiman1996bagging}. The aggregation averages over the versions and does a popularity vote when predicting a class. The multiple versions are formed by making bootstrap replicates of the learning set $D^o$ and new arriving dataset $D^n$, and using $D=D^o+D^n$ as new learning sets, which is represented by $D_i (i \in \left\{1,2,...,s\right\})$. The accuracy of each tree is measured by out-of-bag error rates, whose distribution for all the trees is $O$ and out-of-bag error rate for each tree is $o_i  (i \in \left\{1,2,...,s\right\})$. Besides, we establish a standard value $\delta$ to measure the performance of one tree. \par

% Variable description table
\begin{table}[tbp]
\setlength{\abovecaptionskip}{0.cm}
\setlength{\belowcaptionskip}{-0.cm}
\newcommand{\tabincell}[2]{\begin{tabular}{@{}#1@{}}#2\end{tabular}}
\centering
\caption{Variable descriptions}
\label{tab:variable_description}
\begin{tabular}{cc}
\hline
Notation	&Description \\
\hline
$k$	&the number of features \\
$x$	&a feature vector, $x = \left\{x_1,x_2,...,x_k\right\}$ \\
$y,\hat{y}$	&the actual and the predicted label \\
$(x,y)$	&a sample \\
$c,C$	&a single class, the number of classes $c \in \left\{1,2,...,C\right\}$ \\
$s$	&the number of trees in RF \\
$\delta$	&the particular threshold \\
$t$ 	&the computing time \\
$acc$	&Accuracy for RF \\
$T$	&The whole RF of decision tree, $T = \left\{T_1,T_2,...,T_s \right\}$ \\
$O$	&\tabincell{c}{the distribution of out-of-bag error rates,\\ $O=\left\{o_1,o_2,...,o_s\right\}$ } \\
$D,D^o,D^n$	&\tabincell{c}{All, old, new training data, \\ bootstrap sample set is $D_i (i \in {1,2,...,s})$ } \\
\hline
\end{tabular}
\end{table}
% Section OOB estimation
\subsection{OOB estimation}
\label{oob}
OOB estimation is to calculate a threshold $\delta$ to decide if a tree should be retrained or updated only. The threshold can be regarded as a standard representation of an average tree in RF. If the out-of-bag error rate of a single tree is less than $\delta$, it can be regarded as a good tree with pretty good performance and we preserve its structure by updating the leaves node probabilities only, which is called RLP. If not, it will be replaced by a new tree, and the procedure of replacing is called RRN. The whole procedure is presented in Algorithm~\ref{alg:OOB estimation}. \par

% Algorithm 2:OOB estimation
\begin{algorithm}[htb]
\centering
\caption{OOB estimation}
\label{alg:OOB estimation}
\begin{algorithmic}[1]
\REQUIRE~~
\\
The whole RF, $T$; \\
Bootstrap samples for each tree, $D_i$ ($i \in {1,2,...,s}$); \\
Old training data, $D_o$; \\
New training data, $D_n$; \\
\ENSURE~~
\\
The threshold, $\delta$;
\FOR{each tree $T_i$ in $T$}
	\STATE $D=D^o+D^n$
	\STATE $D^l=D-D_i$    \hspace{3cm}// $D^l$:left-out sample set
	\FOR{($x,y$) in $D^l$}
		\STATE $\hat{y}\leftarrow T_i(x)$
		\IF{$\hat{y}=y$}
		\STATE $I\left\{\hat{y}=y\right\}=1$ \hspace{3cm}// $I\left\{\hat{y}=y\right\}$:loss function
		\ELSE
		\STATE $I\left\{\hat{y}=y\right\}=0$
		\ENDIF
	\ENDFOR
	\STATE $o_i=\frac{\begin{smallmatrix} \sum_{(x,y) \in D^l} I\left\{\hat{y}=y\right\}\end{smallmatrix}}{\left|D^l\right|}$ \hspace{2cm}// calculate the out-of-bag error for $T_i$
\ENDFOR
\STATE $O\sim N(\mu,\sigma^2)$
\STATE $(\mu,\sigma^2)\leftarrow MaxLikelihoodEstimation(O,\mu,\sigma^2)$
\STATE $\delta=\mu$
\RETURN $\delta$
%\STATE $\delta=\frac{\begin{matrix}\sum_{i=1}^so_i\end{matrix}}{s}$ \hspace{3cm}// calculate $\delta$
\end{algorithmic}
\end{algorithm}
The most important thing in OOB estimation is to calculate the out-of-bag error of each tree. Each tree is constructed using bootstrap samples from training set, so the left-out samples can be used to measure the performance of it~\citep{breiman1996out}. Therefore, the out-of-bag error rate of each tree can be computed, and the particular threshold can be estimated according to the mean value of out-of-bag error's Gaussian distribution. We will show the computing procedure in detail as following. \par
The most important thing is to calculate the out-of-bag error rate for each tree. As mentioned above, $D$ is the whole data set, and $D_i (i\in\left\{1,2,...,s\right\})$ is the bootstrap samples for $T_i$. Thus, the out-of-bag error rate of $T_i$ can be calculated by Equation~\ref{equ:o_i}. \par
% equation
\begin{equation}
\label{equ:o_i}
o_i=\frac{\begin{matrix} \sum_{(x,y) \in D^l} I\left\{\hat{y}=y\right\}\end{matrix}}{\left|D^l\right|}
\end{equation}
\par
$I$ is a loss function, i.e., when a sample is arriving, if the predicted label is the same as actual label, the loss will be $0$, and the loss will be 1 otherwise. The equation can be described by Equation~\ref{equ:loss} blow. \par
% equation
\begin{equation}
\label{equ:loss}
I\left\{\hat{y}=y\right\}=
\begin{cases}
0, &\text{if $\hat{y}=y$} \\
1, &\text{otherwise}
\end{cases}
\end{equation}
\par
Then, achieve the mean value based on the distribution of out-of-bag error. As we know, $o_i (i\in\left\{1,2,...,s\right\})$ are independent and have identical distribution, so the list of them, presented by $O$, is satisfied Gaussian distribution~\citep{hazewinkel2001normal}, i.e., $O\sim N(\mu,\sigma^2)$, in which $\mu$ is the expected value, and $\sigma^2$ is the deviation. Hence, $\delta$ can be represented by the expected value $\mu$, that is, $\delta=\mu$. While $\mu$ can be estimated by $O=\left\{o_1,o_2,...,o_s\right\}$ through maximum likelihood estimation~\citep{aldrich1997ra}. Assuming that the probability of $o \textless o_i$ is $p(o_i)$, we can achieve it by Equation~\ref{equ:p_o}.
% equation
\begin{equation}
\label{equ:p_o}
p(o_i)=\frac{1}{\sqrt{2\sigma^2\pi}} \centerdot e^{- \frac{(o_i-\mu)^2}{2\sigma^2}}
\end{equation}
\par
According to maximum likelihood [reference paper], we should maximize all the probability above, which can be represented by likelihood function~\ref{equ:L}.
% equation
\begin{equation}
\label{equ:L}
L(o_1,o_2,...,o_s;\mu,\sigma)=
\begin{matrix}
\prod_{i=1}^s p(o_i)=\prod_{i=1}^s \frac{1}{\sqrt{2\sigma^2\pi}} \centerdot e^{- \frac{(o_i-\mu)^2}{2\sigma^2}}
\end{matrix}
\end{equation}
\par
Next step, maximizing the $log L(o_1,o_2,...,o_s;\mu,\sigma)$ is to make the gradient of $\mu$ and $\sigma$ equal to zero. With these, we can calculate the threshold $\delta$ throughout $\mu$. \par
However, the data will be rollingly released more than once, so keep the stability of the model is very important. It is convinced that the trees updated by RLP is not that reliable, because this approach doesnot change the splitting function or size of the trees. Therefore, when forest need to be updated the second time or more, the trees are retrained by RRN or RLP will be well balanced. To make a balance, OOB boosting is proposed to do that. It boosts the out-of-bag-error rate of trees updated by RRN, which helps us to retrain the tree updated by RLP compared the one updated by RRN with almost the same accuracy. \par
The boosting out-of-bag-error rate $o$ is decided by a learning rate $\alpha$ and $tanh(o)$, which is depicted in Equation~\ref{equ:boost}.

% equation
\begin{equation}
\label{equ:boost}
o=o+\alpha*tanh(o)
\end{equation}

% section RRN
\subsection{RRN}
% the body for RRN
Rolling release NCM decision trees (RRN) presented in this paper is a rolling method to incrementally incorporate new classes of data unlimitedly. It modifies the current classifier to a new one with trees re-constructed by learning new bootstrapping samples when new data arrives. RRN is described in Algorithm~\ref{alg:RRN} and will be explained in detail as following. \par

% Algorithm 3:RRN
\begin{algorithm}[htb]
\centering
\caption{Rolling release NCM decision tree(RRN)}
\label{alg:RRN}
\begin{algorithmic}[1]
\REQUIRE~~
\\
The previous random forest, $T^o$; \\
The out-of-bag error for each tree, $o_i$;\\
The threshold, $\delta$; \\
All training data, $D$; \\
\ENSURE~~
\\
The new random forest, $T^n$;

\FOR{each tree $T_i$ in $T^o$}
	\IF{$o_i > \delta$}
	\STATE \textbf{$Discarding(T_i)$}
	\STATE $D_i\leftarrow Bootstrap(D)$
	\STATE // The growing a new NCM decision tree
	\STATE \textbf{$T_i\leftarrow Growing(D_i):$}
	\IF{all the $(x,y)$ in $D_i$ has the same label $k$ or reach the max Depth}
		\RETURN classes probabilities $P$
	\ELSE
		\STATE // $K$ is a random subset of the classes observed in $D_n$
		\STATE $K \leftarrow ClassesSubset(D_i)$
		\STATE class centroids $\theta_n \leftarrow$ \textbf{$CalClassCentroids(D_n)$}
		\STATE $K_{left},K_{right} \leftarrow$ \textbf{$ChooseBestFeature(D,\theta_n)$} according to Information gain
		\STATE $D_{left},D_{right} \leftarrow$ \textbf{$SplitDataSet(D,K_{left},K_{right},\theta_n)$}
		\STATE build subtree:$T_{left}=Growing(D_{left}),T_{right}=Growing(D_{right})$
		\STATE $T \leftarrow K_{left}:T_{left}+K_{right}:T_{right}$
	\ENDIF
	\RETURN $T$
	\ENDIF
\ENDFOR
\STATE $T^n \leftarrow \left\{T_1,T_2,...,T_s\right\}$
\end{algorithmic}
\end{algorithm}

% RRN body
Each time new data arrives, we calculate a threshold according to the OOB estimation. For each tree, if its OOB error is less than the threshold, it means this tree does not reach the average level. Thus, it will be discarded and replaced by a new NCM decision tree~\citep{ristin2014incremental}, \citep{ristin2016incremental}. The training samples of a new tree are bootstrapping samples from the new and old data. When all the trees were judged, RRN outputs a new bagging forest to make data classification. The growing of a NCM decision tree is illustrated in function $Growing$. \par

% growing new decision tree body
The construction of constructing a NCM decision tree is an iterative process. Once all the samples in a node share the same class or this tree has reach the max depth, the algorithm will return the leaves node probabilities of classes. Each iterative procedure consists of $3$ functions, they will be described in detail as follows. \\
1) CalClassCentroids \par
Since performance of a decision tree is heavily depended on splitting function, NCM decision trees take class centroid instead of best feature. Supposing samples $D_n$ arrive at node $n$, we can randomly select a subset $K$ of classes observed in $D_n$. For each class $k$ in $K$, $D_n^k$ is the subset of $D_n$ of class $k \in K$ and the corresponding class centroid $\theta_n^k$ can be calculated by Equation~\ref{equ:theta_n^k}.
% equation
\begin{equation}
\label{equ:theta_n^k}
\theta_n^k=\frac{1}{|D_n^k|}\centerdot\begin{matrix}\sum_{i \in D_n^k}x_i\end{matrix}
\end{equation}
\\
2) ChooseBestFeature \par
After calculating the class centroids in node $n$, they are randomly assigned to left or right child node and form two class sets $K_{left}$ and $K_{right}$. This is the split function of data, which is represented by $f_n:x \rightarrow \left\{0,1\right\}$. According to it, $D_n$ can be divided into $D_{left}$ and $D_{right}$. In this paper, information gain is taken to measure the split. First, information entropy of $D_n,D_{left}$ and $D_{right}$ are computed as $E_n$, $E_{left}$ and $E_{right}$. Second, calculate the information gain described in Equation~\ref{equ:gain}.

\begin{equation}
\label{equ:gain}
Gain(D_n,f)=E_n-\begin{matrix}\sum_{i \in \left\{left,right\right\}}E_i\end{matrix}
\end{equation}
\par
Assuming the random splitting function space is $F_n$, so we can choose the best one with the most information gain as Equation~\ref{equ:bestF}.

\begin{equation}
\label{equ:bestF}
f_n=\underset{f \in F_n}{\mathrm{argmax}}Gain(D_n,f)
\end{equation}
\par

From mentioned above, a few comparisons is needed at each node, thus reducing the time cost to resort to expensively learn best feature and offering non-linear classification at node level. While that is important to large-scale data classification, and the performance of NCM classifier is comparable to that of linear SVMs which obtain current state-of-the-art performance~\citep{mensink2013distance}, so we take use of NCM classifier instead of decision trees. \\
3) SplitDataSet \par
As we choose the best split function, the class centroids can be distributed to the left or right node. Thus, we can split the data into two datasets according to the distance between data and class centroids, which is defined by Equation~\ref{equ:label}.
% equation
\begin{equation}
\label{equ:label}
\hat{y}=\underset{k \in K}{\mathrm{argmin}}||x-\theta_n^k||^2
\end{equation}
\par

To meet the requirement of constantly growing datasets, we have to update the trees continuously, which is similar with rolling release. For rolling release NCM decision trees, there are two ways to explain rolling release. First, when data is arriving, in the procedure of bagging, we judge all the trees and update one by one. Second, data growing will not stop, so each time data rolling falls, we rolling release the whole forest. It is also another rolling release. \par

As we all know, trees are generated by training samples, if you want to change the data, you must update the tree. Once concerned with extending, cutting, modifying any sub-tree of the previous tree, data category and distribution will be the bottleneck. To avoid that, we just grow a new entire tree without influencing the trees before. Also, the training data is the bootstrapping dataset of the original and present data, it makes our models suit to incorporate new classes of data unlimitedly. \par

% section RLP
\subsection{RLP}

ReGenerate leaves probabilities (RLP) is a light weight method to update the probabilities of the leaf node without structure modification of a decision tree, which contributes a lot to computational cost reduction. The whole procedure is described in Algorithm~\ref{alg:RLP} and will be explained in detail as following. \par
% Algorithm 3:RRN
\begin{algorithm}[htb]
\centering
\caption{ReGenerate leaves probabilities (RLP)}
\label{alg:RLP}
\begin{algorithmic}[1]
\REQUIRE~~
\\
The previous random forest, $T^o$; \\
The out-of-bag error for each tree, $o_i$;\\
The threshold, $\delta$; \\
All training data, $D$; \\
\ENSURE~~
\\
The new random forest, $T^n$;

\FOR{each tree $T_i$ in $T^o$}
	\IF{$o_i <= \delta$}
		\STATE $Dismiss(T_i, leaves probabilities)$
		\STATE $D_i \leftarrow Bootstrap(D)$
		\STATE $K \leftarrow ClassesSet(D_i)$
		\STATE // The updating of a decision tree
		\STATE \textbf{Define $T_i \leftarrow Updating(T_i,D_i):$}
		\STATE $n\_node \leftarrow NumberOfLeavesNode(T_i)$
		\STATE \textbf{$\left\{D_i^1,D_i^2,...,D_i^{n\_node} \right\} \leftarrow Fall(T_i,D_i)$}
		\FOR{$D_i^j \in \left\{D_i^1,D_i^2,...,D_i^{|K|} \right\}$}
			\FOR{$k$ in $K$}
				\STATE $D_i^j(k) \leftarrow DatasetOfClassK(D_i^j,k)$
				\STATE $P_i^j(k)=\frac{|D_i^j(k)|}{|D_i^j|}$
			\ENDFOR
			\STATE $P_i^j=\left\{P_i^j(1),P_i^j(2),...,P_i^j(|K|)\right\}$
		\ENDFOR
		\STATE $P_i=\left\{P_i^1,P_i^2,...,P_i^{n\_node}\right\}^\mathrm{T}$
		\STATE update $P_i$ to $T_i$
		\STATE \textbf{EndDefine}
	\ENDIF
	\STATE $T^n \leftarrow \left\{T_1,T_2,...,T_s\right\}$
\ENDFOR

\end{algorithmic}
\end{algorithm}

RLP also takes use of the threshold generated by OOB estimation. For each tree in RF, if the out-of-bag error is more than threshold means the versions of trees reach the average level and are robust in learning information. Thus, RLP is to update these trees. Usually, once a decision tree is constructed, the split function in each intern node is determined and the probabilities of leaves is generated. RLP is proposed to update that. \par
Since the splitting function is unchanged, we can put the new data into the tree $T_i$ and get a new data set of class in each leaf node $j$, whose number for a tree is $n\_node$. Supposing the input data is $D_i$ and class set $K$, then the data set of class is $\left\{D_i^j(1),D_i^j(2),...,D_i^j(|K|)\right\}$. Therefore, the probabilities of class $k$ in node $j$ $P_i^j$ can be described in Equation~\ref{equ:prob}.  \par

% equation
\begin{equation}
\label{equ:prob}
P_i^j(k)=\frac{|D_i^j(k)|}{|D_i^j|}
\end{equation}
\par

Thus the probabilities of node $j$ is $P_i^j=\left\{P_i^j(1),P_i^j(2),...,P_i^j(|K|)\right\}$, and we draw a matrix of leaf probabilities of $T_i$ that is $P_i=\left\{P_i^1,P_i^2,...,P_i^{n\_node}\right\}^\mathrm{T}$. Finally, we update the leaf probabilities $P_i$ of $T_i$ and achieve a new tree which can make more classifications without training time apart from traversing the training data once.
\subsection{Computational efficiency}
NCM has been used for large-scale image classification in~\citep{mensink2013distance}, and shown its excellent performance on it. Besides, NCM also reduces time cost when used in RF. As we know, the feature space in a traditional decision tree in RF is $\sqrt{k}$ (It should be $floor(\sqrt{k})+1$, but we leave $floor$ out to make it easy to read), where $k$ is the number of features. However, the NCM decision tree takes centroid centers as split function and it is splitted randomly, so it is a constant time to split the feature space. Therefore, NCM decision tree also outperforms other decision trees like C4.5~\citep{quinlan2014c4} for its time-saving. \par
When new class of data arriving, off-line mode requires us to retrain a new RF, which hinders the computational efficiency greatly. Supposing that, the training time of training a NCM tree is $t_{ncm}$, the whole computation complexity $t_{off}$ for $s$ trees is Equation~\ref{equ:toff}.
% equation
\begin{equation}
\label{equ:toff}
t_{off}=s*t_{ncm}
\end{equation}
\par
While for hi-RF, trees in RF are separated into two groups by the threshold: $T^1=\{T_{11},T_{12},...,T_{1n_1}\}$ (RRN) and $T^2=\{T_{21},T_{22},...,T_{2n_2}\}$ (RLP), and always $n_2<\frac{s}2<n_1$. For trees in $T^1$, they are retrained a new decision tree, and the computational time can be represented by $t_{1}$. For trees in $T^2$, RLP only updates their leaf probabilities, and their computational time can be depicted by $t_{2}$. Therefore, the training time of hi-RF $t_{hi-RF}$ is showed in Equation~\ref{equ:tincre}. \par
% equation
\begin{equation}
\label{equ:tincre}
t_{hi-RF}=n_1*t_{1}+n_2*t_{2}
\end{equation}
\par
While $t_{2}$ is equals to the testing time of all the input data $D$, $t_{1}$ is the training time of $D$, so we can draw a conclusion as described in Equation~\ref{equ:thirf}.
% equation
\begin{equation}
\label{equ:thirf}
t_{hi-RF} \approx n_1*t_{ncm}
\end{equation}
\par
In a conclusion, we can infer that the training time of hi-RF is up to the number of RRN trees. In practice, $n_1$ is always less than $\frac{s}2$, so the training time of hi-RF is much less than off-line random forest.
\section{Experiment}
\label{experiment}
In this section, we take Large Scale Visual Recognition Challenge 2010 (ILSVRC2010) for evaluation~\citep{russakovsky2015imagenet}. It contains 96833 training samples, 15000 test samples, and 15000 validation samples in 100 classes and each sample was represented by 4096 features. We use features extracted by AlexNet~\citep{krizhevsky2012imagenet} in 10 classes. While the data in other categories is not available in public and the feature extracting is beyond the scope of this paper, we will not talk about anything about it. To achieve better performance, the original data are normalized. \par
Since on-line random forest and previous incremental methods either can not handle new classes of data or can just handle one new classes of data, so we cannot make a comparison with them. Therefore, experiments are conducted to compare the accuracy and computational cost of the novel hi-RF with its off-line counterpart on ILSVRC2010. From the result, it demonstrates its suitability on the large scale multi-class incremental data classification. When compared with off-line mode, it only needs much less than the off-line computational with approximately accuracy. Besides, we conduct a serial experiments to verify the stability and effectiveness of hi-RF in the scenario with incrementally increasing data.\par
\subsection{Data pre-processing}
To obtain nice results, data pre-processing is often of vital importance when concerned with exploratory data classification or building a good and robust prediction model. In this paper, we normalized the input data and the accuracy improvement is more than $5\%$. As mentioned in section~\ref{hi-RF}, the centroid of each class $k$ is $\theta_k$ and the deviation can be calculated as $\sigma$. Then, all the samples which can be presented by $(x_1,x_2,…,x_m)$ was normalized by Equation~\ref{equ:norm} below. \par
% equation
\begin{equation}
\label{equ:norm}
x_k=\frac{x_k-\theta_k}{\sigma}
\end{equation}
\par
\subsection{Computational efficiency and Accuracy}
\begin{figure}
\setlength{\abovecaptionskip}{0.cm}
\setlength{\belowcaptionskip}{-0.cm}
\centering
\subfigure{
\label{fig:subfig:accuracy} %% 第一幅图的标签
\includegraphics[width=0.3\textwidth]{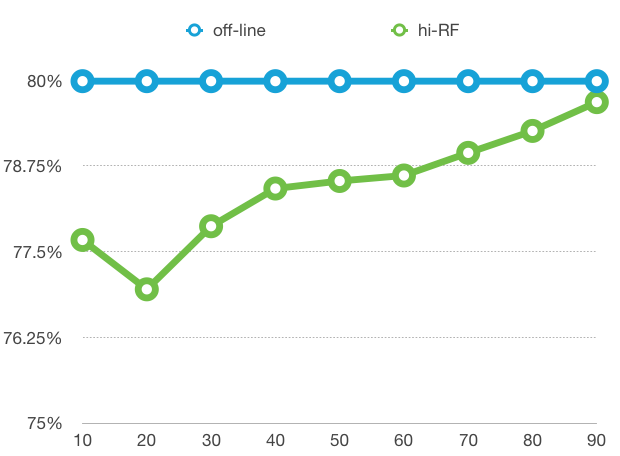}}
\subfigure{
\label{fig:subfig:training_time} %% 第二幅图的标签
\includegraphics[width=0.3\textwidth]{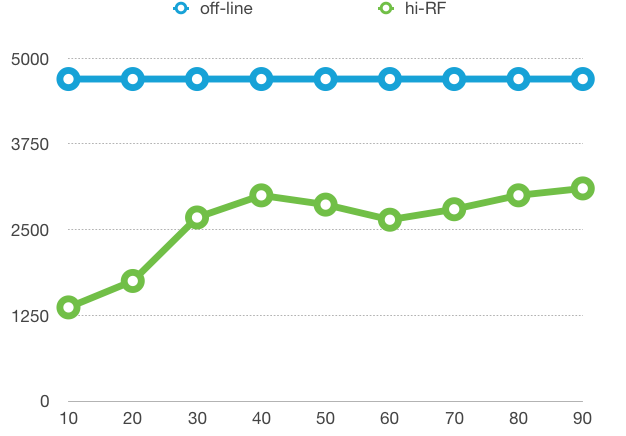}}
\subfigure{
\label{fig:subfig:testing_time} %% 第二幅图的标签
\includegraphics[width=0.3\textwidth]{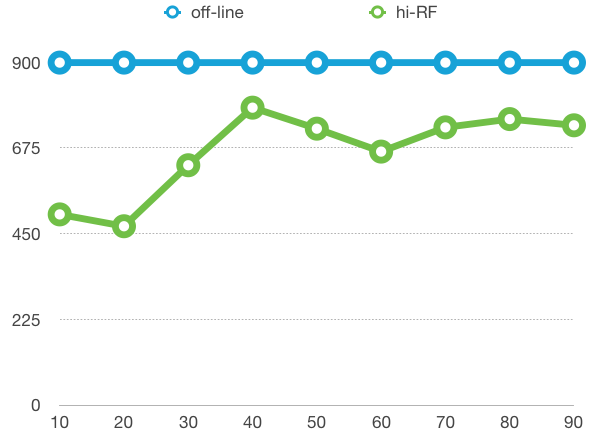}}
\caption{Comparison between baseline and hi-RF, while the original class number is range from $10$ to $90$, and the added data range from $90$ to $10$, and the final data class number is $100$ a) Accuracy b) Training time c) Testing time}
\vspace{-1cm}
\label{fig:e1} %% label for entire figure
\end{figure}

In this section, computational efficiency and accuracy are evaluated in nine different proportions of the class number of incremental data and the original data. We cannot even try all kinds of proportions, so we take the above nine as representations, and we take $50$ trees constructing the RF. After the procedure hi-RF, the nine situations have $100$ classes data, while the original data is $n$, $n\in\{10,20,30,40,50,60,70,80,90\}$. Accuracy, training time and testing time are compared between hi-RF and off-line method, which is the baseline, while hi-RF has $n$ classes of original data and off-line random forest retrain $50$ trees with all the data. The final results are showed in Figure~\ref{fig:e1}. \par
% figure
\textbf{Accuracy.} Figure~\ref{fig:subfig:accuracy} plots the accuracy for the baseline and our approach. It shows that when the original data is small, the accuracy of hi-RF is not that satisfied to baseline. Especially when the original data is $20$, the accuracy of final 100 classes data is down to the mountain valley $76.95\%$. Nonetheless, after that, the accuracy is increased with the number of the initial classes. It ranges from $76.95\%$ to $79.43\%$, while the accuracy of off-line forest is $79.65\%$. The class number of original data is closer to all the data after added, the result is more accurate. We can also infer that the difference ranges from $0.22\%$ to $2.70\%$, thus showing the stability regardless of the original class number. \par
\textbf{Training time.} Figure~\ref{fig:subfig:training_time} plots the training time for the baseline and our approach. The least training time is $1365.6$, and the most one is $2396.12$, while the off-line computational time is $3733.23$. Although the training time is theoretically half the baseline, the complexity of the tree's structure for updating the leaf node probabilities creates an additional overhead. When the structure of trees are simple just as initial number is less than $40$, the training time is less than half off-line computational time, and more than half otherwise. \par
\textbf{Testing time.} Figure~\ref{fig:subfig:testing_time} plots the testing time for the baseline and our approach. The final testing time is varied with the trees complexity, the curve increases for the trees complexity is increases sharply and gently increases for the trees' structure is stable. \par
Finally, we report the accuracy, training time and testing time of hi-RF and baseline on $100$ classes incrementally added from $n$, $n\in\{10,20,30,40,50,60,70,80,90\}$. From massive experiments, it is proved practically that hi-RF requires more a little more than half computational time of the baseline mode with negligible loss of accuracy. It is not sensitive to the original classes.
\subsection{Stability}
\begin{figure}
\setlength{\abovecaptionskip}{0pt}%
\setlength{\belowcaptionskip}{0pt}
\centering
\subfigure{
\label{fig:subfig:e2_1_accuracy} %% 第一幅图的标签
\includegraphics[width=0.3\textwidth]{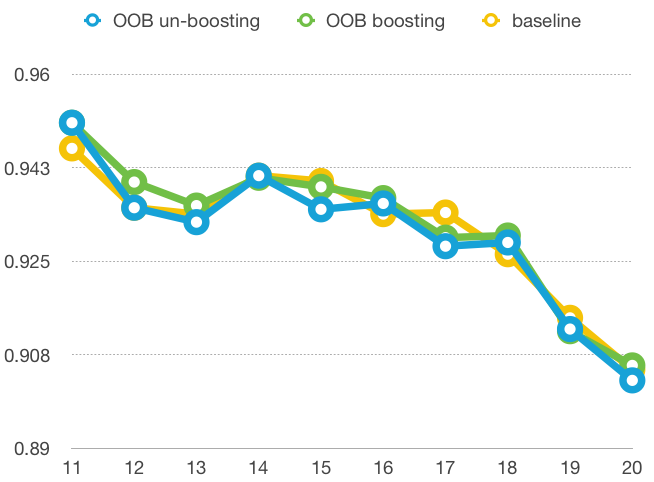}}
\subfigure{
\label{fig:subfig:e2_1_training_time} %% 第二幅图的标签
\includegraphics[width=0.3\textwidth]{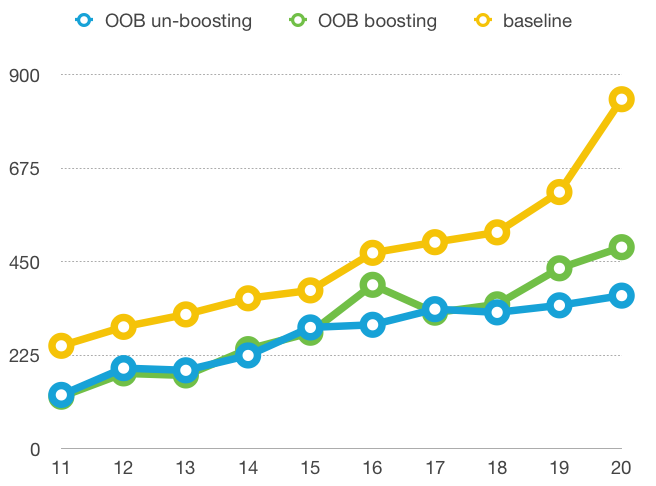}}
\subfigure{
\label{fig:subfig:e2_1_testing_time} %% 第二幅图的标签
\includegraphics[width=0.3\textwidth]{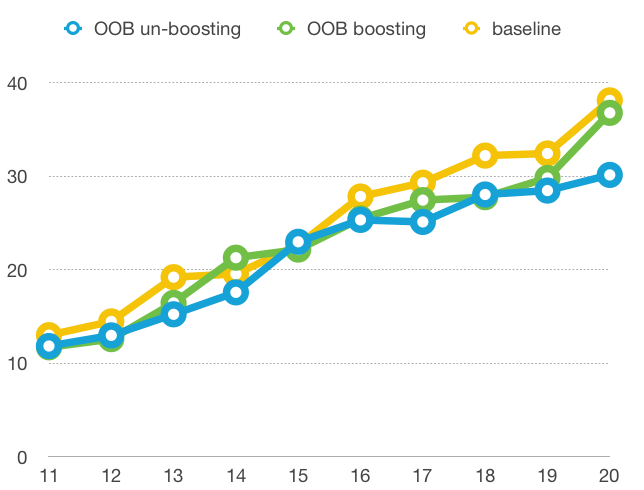}}
\caption{Comparison among baseline, hi-RF with OOB boosting and OOB unboosting, while the original class number is range from $10$ to $20$, and the step size is $1$ a) Accuracy b) Training time c) Testing time}
\label{fig:e2_1} %% label for entire figure
\vspace{-1cm}
\end{figure}

\begin{figure}[htp]
\setlength{\abovecaptionskip}{0pt}%
\setlength{\belowcaptionskip}{0pt}
\centering
\subfigure{
\label{fig:subfig:e2_5_accuracy} %% 第一幅图的标签
\includegraphics[width=0.3\textwidth]{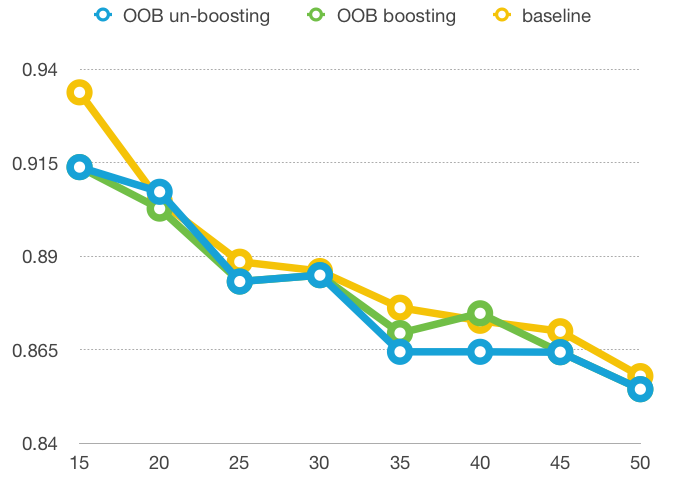}}
\subfigure{
\label{fig:subfig:e2_5_training_time} %% 第二幅图的标签
\includegraphics[width=0.3\textwidth]{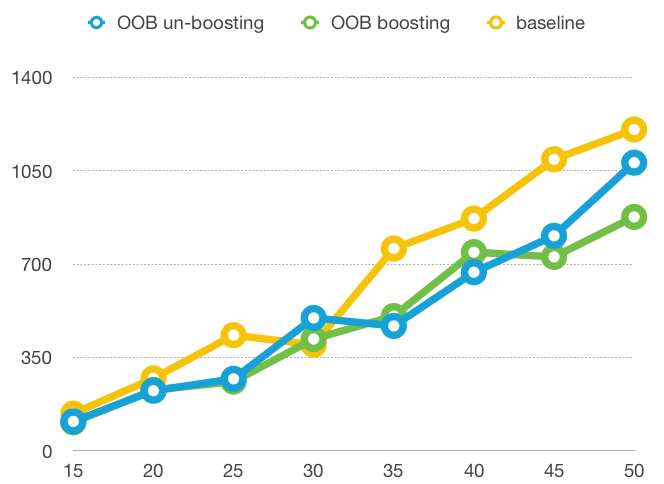}}
\subfigure{
\label{fig:subfig:e2_5_testing_time} %% 第二幅图的标签
\includegraphics[width=0.3\textwidth]{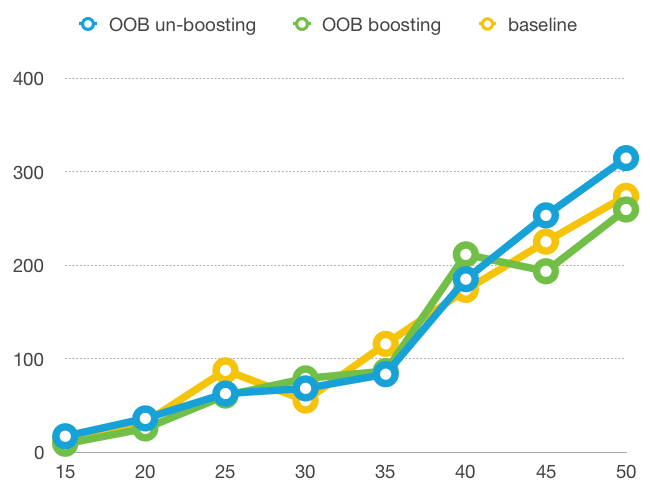}}
\caption{Comparison among baseline, hi-RF with OOB boosting and OOB unboosting, while the original class number is range from $10$ to $50$, and the step size is $5$ a) Accuracy b) Training time c) Testing time}
\label{fig:e2_2} %% label for entire figure
\vspace{-1cm}
\end{figure}

\begin{figure}[htp]
\setlength{\abovecaptionskip}{0pt}%
\setlength{\belowcaptionskip}{0pt}
\centering
\subfigure{
\label{fig:subfig:e2_10_accuracy} %% 第一幅图的标签
\includegraphics[width=0.3\textwidth]{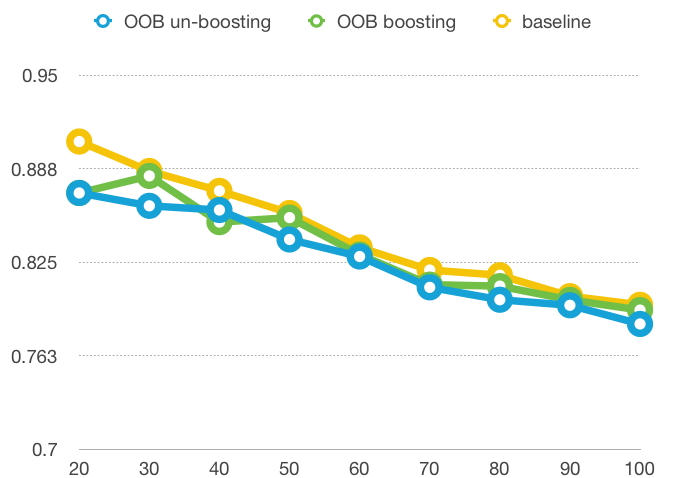}}
\subfigure{
\label{fig:subfig:e2_10_training_time} %% 第二幅图的标签
\includegraphics[width=0.3\textwidth]{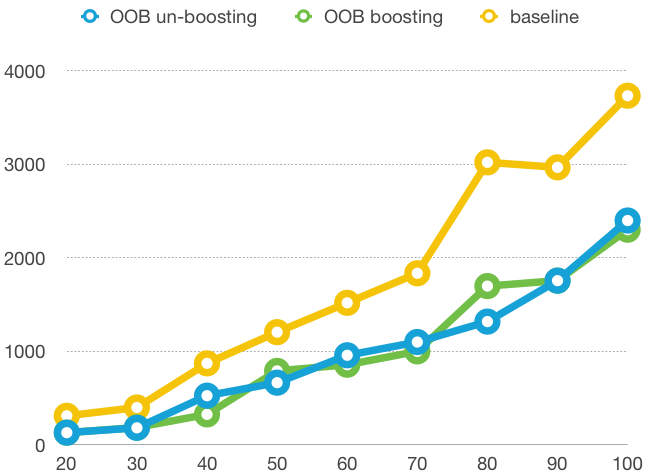}}
\subfigure{
\label{fig:subfig:e2_10_testing_time} %% 第二幅图的标签
\includegraphics[width=0.3\textwidth]{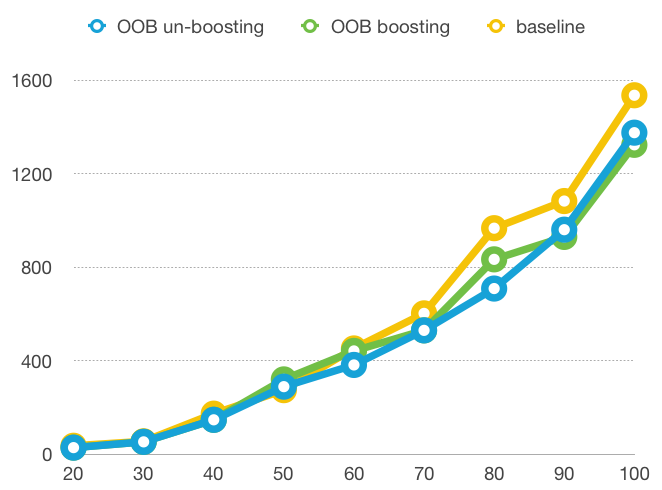}}
\caption{Comparison among baseline, hi-RF with OOB boosting and OOB unboosting, while the original class number is range from $10$ to $100$, and the step size is $10$ a) Accuracy b) Training time c) Testing time}
\label{fig:e2_3} %% label for entire figure
\vspace{-1cm}
\end{figure}

In practices, multiple classes can appear in batches, so to verify the stability in different senario, the number of simultaneous classes to add is an interesting parameter to study. According to the sparse batch and intensive batch, the experiment designs to three part with different incremental strategy, i.e., different batch size. The initial random forest was constructed by $10$ classes of training data, and new arriving data is increased in batch. The batch size in the three group of experiments are $1$, $5$, $10$, and the number of classes of final data is $20$, $50$, $100$. Experiments were conducted between hi-RF and off-line random forest as baseline, especially, hi-RF with using OOB boosting and without using OOB boosting were also compared to have a better understanding of its stability when applying OOB boosting to it, which is showed in~\ref{fig:e2_1},~\ref{fig:e2_2},~\ref{fig:e2_3}.
% figure
% figure
% figure
\\
1) Step size equals to $1$ \par
Comparison with step size $1$ among baseline, hi-RF with OOB boosting and OOB unboosting is showed in Figure~\ref{fig:e2_1}. The data increases so slow that the hi-RF with OOB boosting or OOB un-boosting accuracy keeps pace with the off-line mode. While the training time and testing time curve trend keeps similar with~\ref{fig:subfig:training_time} and~\ref{fig:subfig:testing_time}.\\
2) Step size equals to $5$ \par
Figure~\ref{fig:e2_2} shows the above three comparisons. The curve trend is also the same with~\ref{fig:e2_1}. So we will not show it in detail.\\
\\
3) Step size equals to $10$ \par
Figure~\ref{fig:e2_3} shows the above three comparisons. It said that, hi-RF is very stable, although the step size is big. OOB boosting works a little to the hi-RF without OOB boosting, because the accuracy in green is always higher than the curve in blue. \par
In a conclusion, the accuracy loss of hi-RF with OOB boosting and without OOB boosting keep pace with that in last section, and shows that the stability even the initial model is training by few classes of data. Besides, it seems that hi-RF with OOB boosting applied is a little more stable and accurate than hi-RF without OOB boosting applied.
\section{Conclusion}
\label{conclusion}
In this paper, we describe a hi-RF which integrates new classes gracefully for large-scale multi-class data classification. Extensive experiments were performed and showed that it preserved the overall accuracy with much less computational cost. Therefore, we can implement this method to scenario when few training samples are available at the beginning, and it can improve performance with least cost. \par
Each tree in a forest is built and tested independently from other trees. Hence the overall training and testing procedures can be performed in parallel later. Although the feature space is large, only a fraction of it will actually useful. Therefore, we will design a paralleled algorithm to make use of the informative features only.
\bibliography{acml16}
\end{document}